\documentclass[letterpaper, 10 pt, conference]{ieeeconf}  

\IEEEoverridecommandlockouts                              
\usepackage{graphicx}
\usepackage{epsfig}
\usepackage{float}
\usepackage{threeparttable}
\usepackage{subfigure}
\usepackage{url}
\usepackage{amsmath} 
\usepackage{amssymb}  
\usepackage{color}
\usepackage{xcolor}
\usepackage{float} 

\definecolor{skyblue}{rgb}{0.529, 0.808, 0.922} 
\definecolor{lightblue}{rgb}{0.678, 0.847, 0.902}
\usepackage[ruled,vlined]{algorithm2e}
\usepackage{algpseudocode}
\usepackage{gensymb}
\usepackage{bigstrut,multirow}
	
\usepackage[colorlinks=false, pdfborder={0 0 0}]{hyperref}
\usepackage{cleveref}
\usepackage{booktabs} 
\usepackage{multirow}
\usepackage{graphicx}
\usepackage{makecell}

\usepackage{cite} 

\crefname{figure}{fig.}{figs.}
\crefname{table}{tab.}{tabs.}
\crefname{equation}{eq.}{eqs.}
\usepackage{microtype}

\title{\LARGE \bf
ArtiSG: Functional 3D Scene Graph Construction via Human-demonstrated Articulated Objects Manipulation
}

\author{Qiuyi Gu$^{*}$, Yuze Sheng$^{*}$, Jincheng Yu, Jiahao Tang, Xiaolong Shan, Zhaoyang Shen, Tinghao Yi,\\
Xiaodan Liang, Xinlei Chen, Yu Wang  
\thanks{Qiuyi Gu, Yuze Sheng, Jincheng Yu, Jiahao Tang, Xiaolong Shan, Zhaoyang Shen, Xinlei Chen, and Yu Wang are with the Tsinghua University, China. Tinghao Yi is with the University of Science and Technology of China and Openmind, China. Xiaodan Liang and Qiuyi Gu are with the Pengcheng Laboratory, China.}%
\thanks{$^{*}$ Contributed equally to this work.}%
}

\begin{document}

\maketitle
\thispagestyle{empty}
\pagestyle{empty}

\begin{abstract}
3D scene graphs have empowered robots with semantic understanding for navigation and planning.
\textcolor{black}{However, current functional scene graphs primarily focus on static element detection, lacking the actionable kinematic information required for physical manipulation,} particularly regarding articulated objects.
Existing approaches for inferring articulation mechanisms from static observations are prone to visual ambiguity, while methods that estimate parameters from state changes typically rely on constrained settings such as fixed cameras and unobstructed views.
Furthermore, \textcolor{black}{inconspicuous functional elements like hidden handles are frequently missed by pure visual perception}. 
To bridge this gap, we present ArtiSG, a framework that constructs functional 3D scene graphs by encoding human demonstrations into structured robotic memory.
Our approach leverages a robust data collection pipeline utilizing a portable hardware setup to accurately track 6-DoF manipulation trajectories and estimate articulation axes, even under camera ego-motion.
By integrating these kinematic priors into a hierarchical, open-vocabulary graph, \textcolor{black}{our system not only models \textit{how} articulated objects move but also utilizes physical} interaction data to discover implicit elements.
Extensive real-world experiments demonstrate that ArtiSG significantly outperforms baselines in functional element recall and articulation estimation precision. 
Moreover, we show that the constructed graph serves as a reliable robotic memory, effectively guiding robots to perform language-directed manipulation tasks in real-world environments containing diverse articulated objects.
\end{abstract}

\section{Introduction}
\label{sec:intro}
Scene understanding is fundamental for robots operating in complex and unstructured environments.
Recent research on 3D scene graphs has made significant progress in semantic understanding, enabling applications such as language-guided object retrieval \cite{11246569, gu2025mr}, navigation \cite{werby23hovsg, chen2023not}, and planning \cite{gu2024conceptgraphs, rana2023sayplan}.
However, real-world manipulation requires robots to go beyond mere semantic categorization and master the physical properties of their surroundings, particularly those of functionally intricate articulated objects \cite{Hsu2023DittoITH}.
This functional awareness is essential to bridge perception with action, facilitating physically grounded and task-aware interactions in human-centric environments. 
Motivated by this necessity, our work aims to augment 3D scene graphs with functional information derived from articulated objects.

Understanding object articulation remains a longstanding challenge, primarily due to the vast diversity in visual appearance and internal kinematic mechanisms. 
Recent data-driven approaches \cite{wu2022vatmart, zhang2025adaptive, yuan2025general} have attempted to infer articulation trajectories directly from static visual observations.
However, these methods often struggle with visual ambiguities \cite{li2025flowbothd}, where objects with distinct mechanisms share highly similar appearances.
Another line of research \cite{jiang2022ditto, jiayi2023paris, liu2025building} estimates axes of articulated objects by observing state changes before and after manipulation. 
Yet, these approaches typically rely on constrained settings, such as unobstructed views and fixed camera perspectives, which are difficult to guarantee in unconstrained real-world scenarios.

A further challenge lies in guiding robots to perform effective interactions with articulated objects in complex scenes.
\textcolor{black}{Recent pioneering works \cite{rotondi2025fungraph, zhang2025open,corsetti2025fun3du} have made significant strides by introducing fine-grained functional element detection into 3D scene representations. 
However, they remain limited to static semantic awareness. 
They identify \textit{where} the functional elements are, but fundamentally fail to model \textit{how} they move.} 
Furthermore, \textcolor{black}{inconspicuous elements without explicit visual features are frequently missed by pure visual perception.}

\begin{figure}[t!]
    \centering
    \includegraphics[width=1\linewidth]{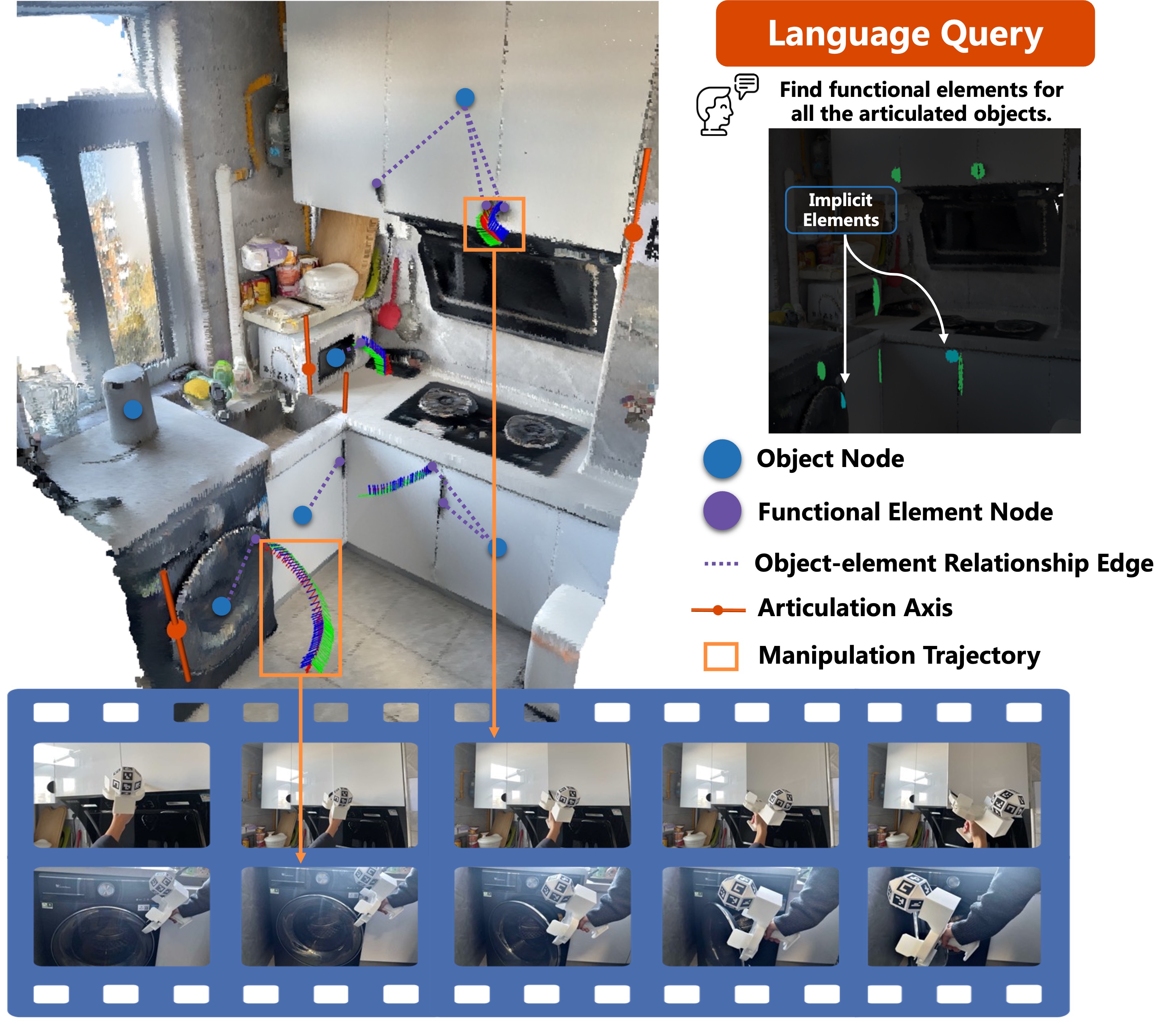}
    \caption{\textbf{Constructing Functional Scene Graphs via Human Demonstration.} The bottom film strips show our manipulation sequences using a custom UMI gripper. From these sequences, we extract articulation trajectories and estimate axes, registering them to the corresponding element nodes in the graph. This structured representation enables open-vocabulary queries to locate functional elements and provides actionable priors for robot manipulation.} 
    \label{fig:sysframe}
    \vspace{-4mm}
\end{figure}

To tackle the challenges above, we draw inspiration from the fact that humans often learn by observing others' manipulations—a capability yet to be fully leveraged for robotic functional scene understanding. 
To bridge this gap, we present ArtiSG, a framework designed to encode observed human manipulations into a structured scene graph, serving as robotic memory to guide subsequent interactions with articulated objects.
As illustrated in \Cref{fig:sysframe}, ArtiSG possesses four key characteristics: 
1) \textbf{a hierarchical graph representation}, which captures the parent-child relationships between objects and their functional elements while allowing various attributes, such as articulation axes and trajectories, to be attached to nodes, 
2) \textbf{viewpoint-robust articulation tracking} that supports dynamic observation perspectives by utilizing a portable setup and gripper pose tracking algorithms to estimate articulation mechanisms from human demonstrations, 
3) \textbf{interaction-augmented functional element detection} through integrating visual foundational models with realistic manipulation trajectories to better identify inconspicuous elements, and 
4) \textbf{open-vocabulary scene construction} where semantic features are aggregated from multiple optimal views for each node via a top-$k$ frame selection mechanism, thereby enhancing generalization and applicability. 


In summary, our contributions are as follows:
\begin{itemize}
\item
We present a \textcolor{black}{systematic framework for constructing functional 3D scene graphs that inherently encode actionable kinematic memory}, jointly capturing both static functional elements and dynamic articulation mechanisms by leveraging vision foundation models and human-demonstrated trajectories.
\item 
We design a viewpoint-robust data collection pipeline utilizing a portable hardware setup to extract articulated object trajectories during human manipulation and accurately estimate articulation axes. 

\item 
We deploy ArtiSG in real-world environments, demonstrating its capability to construct functional 3D scene graphs and its utility in guiding language-based robot manipulation tasks.
\textcolor{black}{The source code and the 3D-printable models for the tracking hardware will be released publicly to lower the deployment barrier and facilitate reproduction.}
\end{itemize}

\section{Related work}
\label{related work}

\subsection{3D Scene Graphs}
Pioneered by Armeni et al. \cite{armeni20193d}, 3D scene graphs abstract environments into nodes and edges, a structure well-suited for encoding attributes and facilitating task planning. 
While many recent works \cite{gu2024conceptgraphs, werby23hovsg, chen2023not, 11246569, gu2025mr, rana2023sayplan} construct object-level scene graphs that support navigation and basic grasping, they often overlook the fine-grained functional details required for articulated object manipulation.
To address this, \textcolor{black}{recent approaches \cite{rotondi2025fungraph, zhang2025open,corsetti2025fun3du} establish a foundation for element-level functional scene understanding. 
For instance, FunGraph \cite{rotondi2025fungraph} trains specialized detectors using custom datasets, while OpenFunGraph \cite{zhang2025open} and Fun3DU \cite{corsetti2025fun3du} leverage vision foundation models for open-vocabulary detection and segmentation. 
These works successfully bridge the semantic gap in static scenes. 
However, because they rely purely on static visual observations, they inherently \textbf{lack the kinematic memory required to model the articulation mechanisms of these elements} during physical manipulation.}

Integrating human interaction into scene construction is an emerging direction. 
Most relevant work to ours is Lost\&Found \cite{behrens2025lost}, which updates the scene graph by tracking human-object interactions. 
It focuses primarily on object tracking, identifying when objects are grasped by humans rather than understanding object kinematics. 
In contrast, our framework treats human interaction as a functional demonstration, leveraging manipulation trajectories to infer and explicitly encode articulation mechanisms into the scene graph.

\subsection{Articulated Object Understanding}
Unlike rigid objects, articulated objects require inferring both actionable parts and kinematic constraints. 
Existing works generally fall into two categories. 
One line of research \cite{wu2022vatmart, zhang2025adaptive, yuan2025general, kuang2025ram} infers contact poses and articulation trajectories directly from static visual observations. 
For instance, GFlow \cite{yuan2025general} and RAM \cite{kuang2025ram} predict motion flows or retrieve trajectories based on large-scale datasets. 
However, relying on static inputs makes these methods \textbf{prone to visual ambiguity}, failing when objects with similar appearances possess distinct internal mechanisms.
Another stream of research \cite{Hsu2023DittoITH, liu2025building, zhang2025iaao} focuses on estimating precise articulation parameters by observing state changes. 
While accurate, these methods typically \textbf{assume constrained settings}, such as fixed camera viewpoints or unobstructed pre- and post-manipulation observations, which are impractical for humans operating in the wild.
\textcolor{black}{Concurrently, ArtiPoint \cite{arti25werby} makes inspiring progress by estimating articulation directly from in-the-wild videos, relaxing the need for fixed viewpoint.}
However, tracking textureless or occluded object parts remains fragile during dynamic interactions.

\begin{figure*}[t]
    \centering
    \includegraphics[width=1\linewidth]{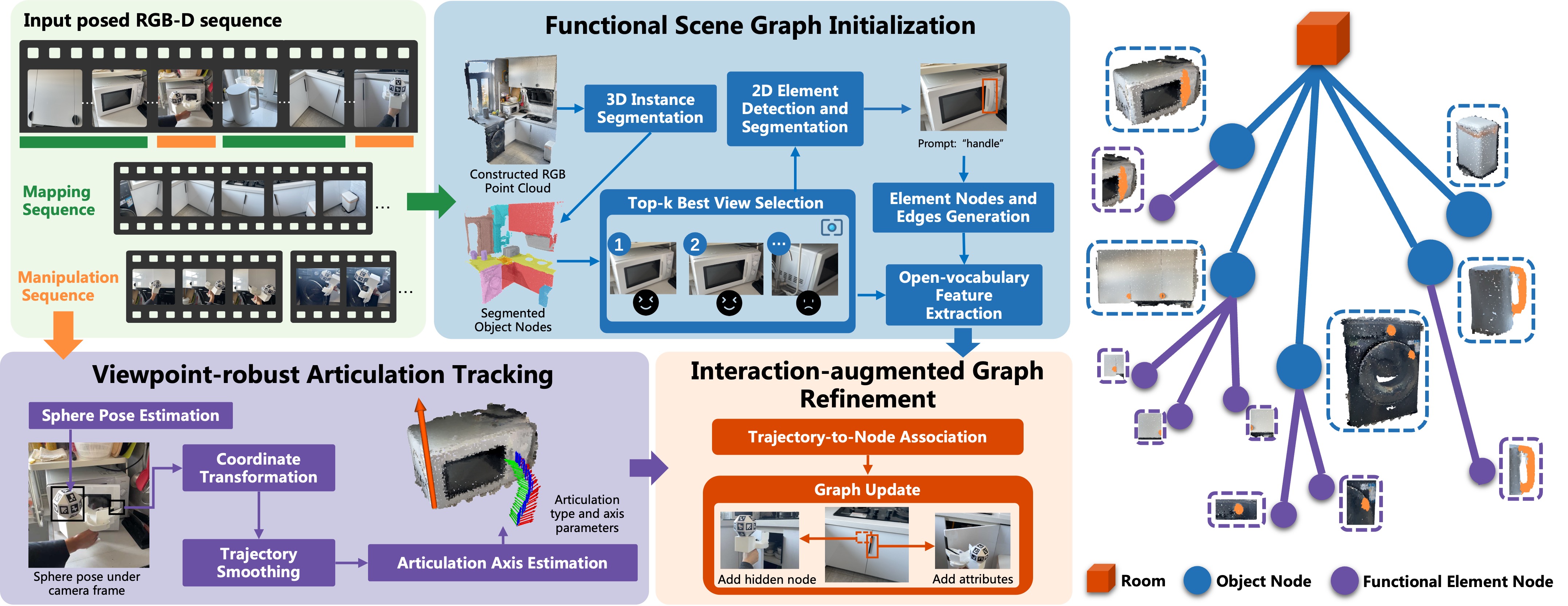}
    \caption{\textbf{System Overview.} Our approach to building the functional scene graph for an indoor room unfolds in three stages. Firstly, the construction begins with the initialization of an element-aware scene representation, where we aggregate multi-view semantics to detect and generate object and functional element nodes that are explicitly visible. Secondly, we leverage a portable setup to track human manipulation, enabling the extraction of precise motion trajectories and the estimation of articulation axes for articulated objects. Finally, we perform interaction-augmented graph refinement, utilizing these human demonstrations to recover inconspicuous functional elements missed in the initial phase and enrich element nodes with articulation kinematic attributes.
    }
    \label{fig:framework}
\end{figure*}

\subsection{Human-demonstrated Manipulation}
Human demonstrations for robots typically stem from in-the-wild videos, teleoperation, or portable interfaces. 
While learning from videos \cite{yuan2025general, kuang2025ram, ju2025robo} offers scalability, it suffers from the embodiment gap and lacks high precision. 
Conversely, teleoperation \cite{wu2025afforddp} ensures high-quality trajectories but relies on specialized hardware, limiting its in-the-wild applicability.

Portable interfaces, such as UMI \cite{chi2024universal} and FastUMI \cite{zhaxizhuoma2025fastumi}, strike a balance by enabling hardware-agnostic data collection in diverse environments.
While these systems primarily utilize wrist-mounted cameras to capture visual data for policy learning, \textbf{our approach adopts a decoupled setup optimized for functional scene understanding}.
We employ a head-mounted camera with built-in SLAM that serves a dual purpose by collecting posed RGB-D frames to construct the static scene graph while tracking the UMI gripper's manipulation trajectories.
This design ensures that both the environmental geometry and the dynamic articulation data are precisely registered within a unified coordinate system, maintaining robustness despite the operator's continuous viewpoint changes.

\section{Problem Formulation}
We aim to construct a functional 3D scene graph for indoor environments populated with articulated objects. 
Formally, we define the scene graph as a tuple $ \mathcal{G} = \{\mathcal{N}^{\mathrm{obj}}, \mathcal{N}^{\mathrm{ele}}, \mathcal{E}\}$.
The set of object nodes $\mathcal{N}^{\mathrm{obj}}$ represents static object bodies.
Each object node $ N_i^{\mathrm{obj}} \in \mathcal{N}^{\mathrm{obj}}$ encapsulates its semantic and geometric attributes, including a category label, an open-vocabulary semantic feature, and the associated point cloud.
The functional element nodes $ \mathcal{N}^{\mathrm{ele}}$ represent actionable components.
Each node $ \mathcal{N}_j^{\mathrm{ele}} \in N^{\mathrm{ele}}$ is characterized by a functional label, an articulation type, an articulation axis $\mathbf{A}_j = \{\mathbf{p}_{c}, \mathbf{p}_{d}\}$ that defines its kinematic mechanism, and a demonstrated manipulation trajectory $\mathcal{T}_j = \{\mathbf{p}_1, \mathbf{p}_2,...,\mathbf{p}_n \}$.
Here, $\mathbf{p}_{c} \in \mathbb{R}^3$ is the center position of the articulation axis and $\mathbf{p}_{d} \in \mathbb{R}^3$ shows the axis's direction.
Each $\mathbf{p}_k \in \mathbb{R}^7$ denotes a 6-DoF pose in the sequence.
The set of edges $\mathcal{E}$ encodes the hierarchical structural relationships, linking a functional element node $ N_j^{\mathrm{ele}}$ to its corresponding parent object node $ N_i^{\mathrm{obj}}$.
This structure supports a one-to-many mapping.
While each functional element belongs to a unique parent object, a single articulated object may possess multiple functional elements.

\section{Approach}
\label{sec:alg}
We propose ArtiSG, a unified framework that constructs functional scene graphs by bridging static visual perception with dynamic human interaction.
Our approach consists of three stages.
Functional Scene Graph Initialization establishes a semantic foundation by aggregating multi-view observations to identify objects and visible functional elements.
Viewpoint-Robust Articulation Estimation leverages a portable interface to capture high-fidelity manipulation trajectories and estimate kinematic parameters.
Interaction-Augmented Graph Refinement fuses these kinematic priors into the graph, explicitly registering articulation attributes and discovering inconspicuous elements missed during the initial visual scan.
\Cref{fig:framework} provides an overview of our approach.

\subsection{Functional Scene Graph Initialization}
\label{sec:initialization}

\noindent\textbf{Object Node Construction}:
We initiate the process by scanning the environment to acquire posed RGB-D mapping sequences and then generating the RGB point cloud of the scene. 
To extract object-level instances, we employ an off-the-shelf 3D instance segmentation model \cite{Schult23ICRA} and then utilize DBSCAN clustering \cite{ester1996density} to remove outliers from each instance. 
The resulting denoised point cloud 
is instantiated as an object node $N_i^{\mathrm{obj}}$ in the graph.

\noindent\textbf{Top-$\boldsymbol{k}$ Frame Selection}:
Detecting fine-grained functional elements and extracting semantic features often rely on 2D detection \cite{liu_grounding_2025, zhou2022detecting} and segmentation models \cite{kirillov2023segment}, as well as vision–language encoders \cite{radford2021learning, Tschannen2025SigLIP2}. 
However, the performance of these models suffers inevitable degradation when target objects are only partially visible or heavily occluded, which is very common when observing elements on articulated objects. 
Therefore, selecting optimal viewpoints that provide sufficient visibility is pivotal. 
For each object node $N_i^{\mathrm{obj}}$, we compute a contribution score $s_{t,i}$ for every frame $t$ in the RGB-D sequence \cite{koch2024open3dsg}. 
Specifically, we project each object's 3D points onto the camera imaging plane using the camera parameters. 
Points falling outside the image boundary or exhibiting significant depth inconsistency which implies occlusion, are filtered out. 
The contribution $s_{t,i}$ is defined as the percentage of valid points retained on the imaging plane relative to the object's total points. 
Based on these scores, we select the top-$k$ frames that offer the most comprehensive observations for each object.

\noindent\textbf{Element Node and Edge Construction}:
Leveraging the selected top-$k$ RGB frames, we proceed to identify functional elements. 
For each frame in top-$k$, we crop the image based on the bounding box of the valid projected points. 
We then employ Grounding DINO \cite{liu_grounding_2025} with predefined prompts (i.e., ``handle'', ``knob'') to detect functional regions, followed by SAM \cite{kirillov2023segment} to obtain fine-grained pixel-level masks. 
These 2D part masks are back-projected into 3D space and observations from multiple views are aggregated into a unified point cloud, resulting in $N_j^{\mathrm{ele}}$.
This multi-view lifting strategy enables us to capture small functional elements that are typically indistinguishable in 3D segmentation. 
Notably, this object-centric processing strategy eliminates the need for a separate edge identification step.
Since functional elements are detected within the visual context of a specific object $N_i^{\mathrm{obj}}$, the belonging edges $E_{ij}$ are naturally established.

\noindent\textbf{Open-vocabulary Feature Extraction}:
Similar to geometric construction, we utilize the cropped top-$k$ frames to compute open-vocabulary features for both object and element nodes. 
We extract features using SigLIP 2 \cite{Tschannen2025SigLIP2} and aggregate them into a single node feature by performing a weighted average based on the frame contribution scores $s_{t,i}$. 
This weighting strategy ensures that views with higher visibility contribute more to the final semantic representation, enhancing robustness against visual ambiguity. 

\subsection{Viewpoint-robust Articulation Estimation}
\label{sec:tracking}
\begin{figure}[t]
    \centering
    \includegraphics[width=1\linewidth]{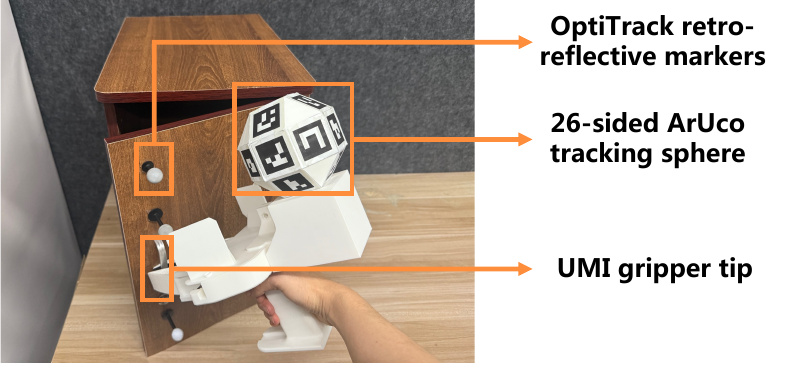}
    \setlength{\abovecaptionskip}{-0.4cm} 
    \setlength{\belowcaptionskip}{-0.4cm}
    \caption{\textbf{Hardware setup for articulation data collection.} The handheld UMI gripper is equipped with a custom 26-sided ArUco tracking sphere, enabling robust 6-DoF pose estimation via a head-mounted camera. OptiTrack retro-reflective markers are attached to the cabinet door to provide ground truth poses for the quantitative evaluation in \Cref{sec: eval_tracking}.}
    \label{fig:UMI}
    \vspace{-4mm}
\end{figure}

\noindent\textbf{Hardware setup}:
Our hardware setup is designed to capture high-fidelity manipulation data despite ego-motion. 
We employ a head-mounted RGB-D camera to visually track a handheld UMI gripper \cite{chi2024universal} fitted with a custom polyhedral sphere.
As shown in \Cref{fig:UMI}, this sphere provides a dense set of ArUco markers that allows the camera to estimate the gripper's 6-DoF pose, while the UMI gripper is a 3D-printed parallel jaw device.
We utilize this rigid gripper interface instead of direct hand tracking for a critical reason.
Human hand-object interactions involve complex and varying contact points, making it difficult to define a consistent reference frame for the object's motion.
In contrast, the UMI gripper acts as a rigid body that stays tightly coupled with the functional element during manipulation.
Therefore, tracking the gripper tip provides a precise proxy for the articulated element's trajectory.
To ensure robustness during mobile operation, the camera utilizes built-in SLAM to establish a globally consistent world frame.
This combination guarantees accurate trajectory recording even when the operator moves freely in the environment.

\noindent\textbf{Trajectory Tracking}:
Our goal is to recover the 6-DoF trajectory of the gripper tip in the world frame, which serves as the demonstrated trajectory $\mathcal{T}_j$ for the functional element node. 
Given RGB-D manipulation sequences from the head-mounted camera, we detect visible ArUco markers on the sphere. 
We establish 2D-3D correspondences by mapping the detected marker IDs $\{u_i^{2\mathrm{D}}\}$ to their pre-calibrated 3D corner positions $\{P_i^{3\mathrm{D}}\}$ on the polyhedral model. 
An initial pose estimate $T_{\mathrm{cam} \leftarrow \mathrm{sphere}} \in SE(3)$ is obtained via a Perspective-n-Point (PnP) solver \cite{bradski2000opencv}:
\begin{equation}
T_{\mathrm{cam} \leftarrow \mathrm{sphere}} = \text{solvePnP}(\{P_i^{3\mathrm{D}}\}, \{u_i^{2\mathrm{D}}\}, K)
\label{equ:pnp}
\end{equation}
where $K$ is the camera intrinsic matrix. 
This local pose is then transformed into the global frame using the real-time camera pose $T_{\mathrm{world} \leftarrow \mathrm{cam}}$:
\begin{equation}
T_{\mathrm{world} \leftarrow \mathrm{sphere}} = T_{\mathrm{world} \leftarrow \mathrm{cam}} \cdot T_{\mathrm{cam} \leftarrow \mathrm{sphere}}
\end{equation}
To suppress jitter caused by hand tremors or detection noise, we process the raw world-frame poses using an adaptive Kalman filter \cite{kalman1960new}. 
Crucially, our filter addresses the cyclic nature of rotation by performing rotation unwrapping, which prevents sudden numerical jumps and ensures smooth angular transitions. 
Furthermore, we adaptively adjust the filter's confidence based on the PnP reprojection error. 
This enables the system to rely heavily on high-quality detections, while automatically prioritizing smooth prediction when markers are partially occluded.
Since the sphere center has a static physical offset with the gripper's tip, we apply a pre-calibrated rigid-body transformation $T_{\mathrm{sphere} \leftarrow \mathrm{tip}}$ to obtain the final end-effector pose:
\begin{equation}
T_{\mathrm{world} \leftarrow \mathrm{tip}} = T_{\mathrm{world} \leftarrow \mathrm{sphere}} \cdot T_{\mathrm{sphere} \leftarrow \mathrm{tip}}
\end{equation}
The resulting sequence forms the final smoothed manipulation trajectory $\mathcal{T}_{j}$.

\noindent\textbf{Articulation Axis Estimation}:
Given the trajectory $\mathcal{T}_j$, we estimate the kinematic mechanism of the object as either a \textit{prismatic} or \textit{revolute} joint using Principal Component Analysis (PCA) and non-linear optimization. 
For a prismatic joint, we apply SVD to the centered points of $\mathcal{T}_j$. The axis direction $\mathbf{p}_d$ is the eigenvector with the largest singular value, and the center $\mathbf{p}_c$ is the trajectory centroid. 
For a revolute joint, we first extract the rotation axis $\mathbf{p}_d$ (the plane normal) as the eigenvector with the smallest singular value. Points are then projected onto the orthogonal plane, where the rotation center $\mathbf{p}_c$ is solved via non-linear least squares to minimize radial deviation. 
The final joint type is selected by comparing the reconstruction residuals of both the prismatic and revolute models. 
\textcolor{black}{To prevent overfitting to the more complex revolute model when a simpler prismatic motion is observed, we utilize the Bayesian Information Criterion (BIC) as the penalty term} for model complexity.
As a result, we obtain the articulation type and axis parameters $\mathbf{A}_j = \{\mathbf{p}_{c}, \mathbf{p}_{d}\}$.

\subsection{Interaction-augmented Graph Refinement}
\label{sec: refine}
\noindent\textbf{Trajectory-to-Node Association} 
\textcolor{black}{To unify the coordinate systems of the mapping sequence in \Cref{sec:initialization} and the manipulation sequences in \Cref{sec:tracking}, the head-mounted camera first scans a fixed ArUco marker in the environment to establish a global anchor frame.}
With trajectories aligned and kinematics estimated, we ground this dynamic data into the static scene graph via geometric matching. 
Specifically, we compute the Euclidean distance between the trajectory's starting pose $\mathbf{p}_1$ (the initial contact point) and the centroids of adjacent functional nodes. 
This distance is evaluated against a spatial threshold to determine if the interaction aligns with a visually detected element or reveals a previously missed one.

\noindent\textbf{Graph Update} 
Based on the association result, we perform either attribute attachment or node instantiation. 
If the nearest node lies within the threshold, we confirm a successful match and register the inferred articulation axis $\mathbf{A}_j$, joint type, and the full trajectory $\mathcal{T}_j$ as dynamic functional attributes of that node. 
Conversely, if no node is found within the threshold, it indicates that the functional element is missed in the initialization step due to occlusion or its implicit nature. 
In this scenario, we instantiate a new functional element node centered at $\mathbf{p}_1$ and explicitly attach the kinematic parameters while linking it to the nearest parent object node. 
This mechanism ensures the scene graph captures a complete set of functional affordances by effectively compensating for visual perception failures through physical interaction.

\section{Experiments}
\label{sec:exp}
In this section, we evaluate ArtiSG in three aspects:
1) scene graph construction quality, assessing the accuracy of functional element detection and open-vocabulary semantic representation;
2) articulation tracking precision, verifying the robustness of our hardware-assisted pipeline against several baselines;
3) real-world manipulation utility, demonstrating the effectiveness of the constructed graph in guiding manipulation tasks.

\subsection{Functional Scene Graph Construction Evaluation}
\label{sec:eval_graph}
\noindent\textbf{Dataset}:
\textcolor{black}{We primarily evaluate our framework in real-world settings, including a kitchen, an office pantry, and a tabletop scene. 
To isolate and validate our static initialization module, we also use three indoor scenes from the Behavior-1k \cite{li2024behavior1k} simulation dataset, as they are densely populated with articulated objects.} 
In total, our evaluation includes 79 articulated objects and 139 functional elements.

\noindent\textbf{Baselines}:
\textcolor{black}{We compare against two SOTA functional scene graph methods FunGraph \cite{rotondi2025fungraph} and OpenFunGraph \cite{zhang2025open}, and a concurrent 3D functional understanding work Fun3DU \cite{corsetti2025fun3du}. 
We also introduce two ablations: \textbf{ArtiSG (w.o human)} relies purely on static initialization without demonstrations, and \textbf{ArtiSG (w.o top-k)} further replaces optimal viewpoint selection with random frames.
Notably, all baselines and ablations utilize only static RGB-D scans, whereas \textbf{ArtiSG (ours)} leverages both static scans and human demonstrations. 
We utilize the prompts ``handle'', ``knob'', and ``button'' in our experiments.} 
Evaluations in \Cref{sec:exp} are conducted on a desktop PC equipped with an Intel I7-13790F CPU and an NVIDIA RTX 4090 GPU.

\noindent\textbf{Metrics}:
\label{sec: metrics}
We evaluate functional element detection using Precision (P), Recall (R), and F1-score, \textcolor{black}{calculated under an IoU threshold of $>0$}. 
We also report the query success rate R@n \cite{zhang2025open}, which measures whether the target object or element is successfully retrieved among the top-$n$ candidates in the scene graph.

\setlength{\tabcolsep}{4pt}
\begin{table}[t!]
\centering
\caption{\textcolor{black}{Functional 3D Scene Graphs Evaluation}}
\begin{tabular}{llccccc}
\toprule
\multirow{2}{*}{\bf{Scene}} & \multirow{2}{*}{\bf{Method}} & \multicolumn{3}{c}{\textbf{Fun. Ele. Node}} & \multicolumn{2}{c}{\textbf{Overall Node}} \\
\cmidrule(lr){3-5} \cmidrule(lr){6-7}
& & $\color{black}\mathbf{R}_{>0}$ & $\color{black}\mathbf{P}_{>0}$ & $\color{black}\mathbf{F1}_{>0}$ & R@1 & R@5 \\

\midrule
\multirow{5}{*}{\makecell[c]{\rotatebox{90}{Real-world}}} 
& OpenFunGraph \cite{zhang2025open}     & 45.7 & 18.4 & 26.3 & 50.0 & 63.1 \\
& \textcolor{black}{FunGraph\cite{rotondi2025fungraph}} & \textcolor{black}{51.9} & \textcolor{black}{23.5} & \textcolor{black}{32.4} & \textcolor{black}{47.6} & \textcolor{black}{64.3} \\

& \color{black} ArtiSG (w.o top-$k$)  & \color{black} 6.2 & \color{black} 5.6 & \color{black} 5.9 & \color{black} 19.0 & \color{black} 21.4 \\ 
& ArtiSG (w.o human) & 55.8 & 41.0 & 47.2 & 60.7 & 70.2 \\
& ArtiSG (ours)      & \bf{88.5} & \bf{51.6} & \bf{65.2} & \bf{79.8} & \bf{89.3} \\

\midrule
\multirow{4}{*}{\makecell[c]{\rotatebox{90}{Simulation}}} 
& OpenFunGraph \cite{zhang2025open}     & 39.7 & 66.1 & 49.6 & 34.0 & 41.8 \\
& \textcolor{black}{FunGraph\cite{rotondi2025fungraph}} & \textcolor{black}{56.1} & \textcolor{black}{\bf{74.3}} & \textcolor{black}{63.9} & \textcolor{black}{44.4} & \textcolor{black}{49.7} \\
& \color{black} ArtiSG (w.o top-$k$) & \color{black} 34.7 & \color{black} 64.8 & \color{black} 45.2 & \color{black} 28.0 & \color{black} 35.3 \\
& ArtiSG (w.o human) & \bf{78.6} & 70.4 & \bf{74.2} & \bf{59.0} & \bf{73.2} \\

\bottomrule
\end{tabular}
\label{table:Scene Graph}
\end{table}

\begin{figure}[t]
    \centering
     \includegraphics[width=1\linewidth]{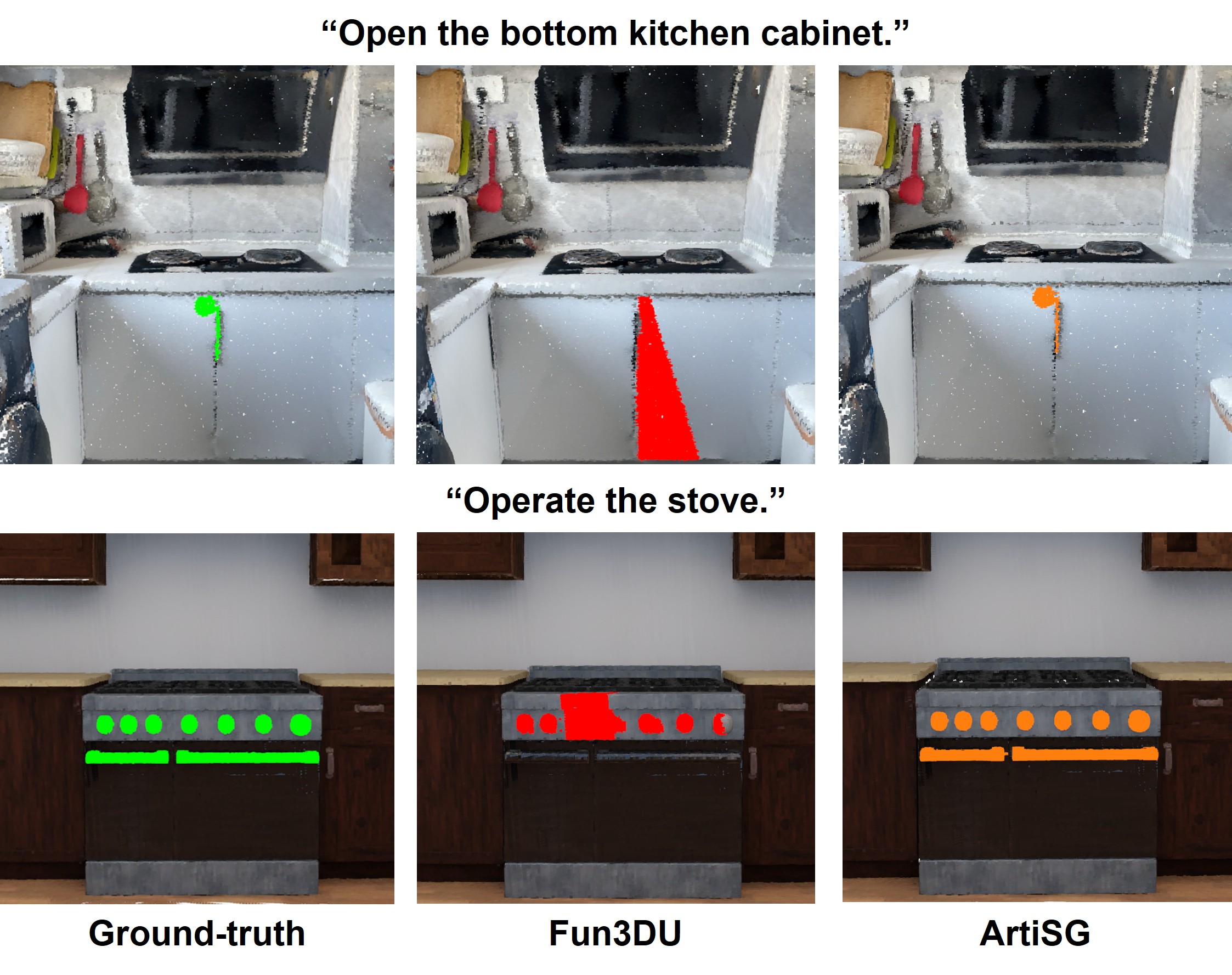}
    \setlength{\abovecaptionskip}{-0.5cm}
    \setlength{\belowcaptionskip}{-0.4cm}
    \caption{\textbf{Qualitative comparison of open-vocabulary querying performance.} We compare query results against \textcolor{black}{Fun3DU. Green masks indicate the ground truth functional elements.}  As shown, our method accurately localizes target elements with high recall \textcolor{black}{and precision}, whereas the baseline often suffer from missed detections or imprecise localization.
    }
    \label{fig:vis_artisg}
    \vspace{-1mm}
\end{figure}

\noindent\textbf{Results on Scene Graph Nodes}:
\textcolor{black}{As shown in \Cref{table:Scene Graph}, ArtiSG demonstrates superior performance in functional element detection and open-vocabulary node retrieval.
For functional element nodes construction, baseline methods rely purely on static features. 
Despite FunGraph's specialized detectors, they both struggle with implicit elements, limiting their real-world recall at 45.7\% and 51.9\%.
Our training-free static initialization module ArtiSG (w.o human) leverages foundation models to achieve comparable performance to FunGraph. 
Crucially, introducing human interaction cues further boosts real-world recall from 55.8\% to 88.5\%, validating that physical interaction effectively uncovers hidden elements.}
Regarding open-vocabulary node retrieval, ArtiSG consistently outperforms baselines in R@1 and R@5 metrics. 
This stems from a larger candidate pool of detected functional element nodes and our top-$k$ frame selection mechanism. 
\textcolor{black}{As evidenced by the significant performance drop in the ArtiSG (w.o top-k) variant across all metrics, aggregating features without viewpoint optimization introduces severe visual noise and occlusions. 
This degradation is particularly drastic in real-world environments due to the vastly larger number of frames per scan, which makes random frame selection highly likely to miss the target elements entirely.
The top-$k$ mechanism effectively mitigates this issue, resulting in far more accurate open-vocabulary representations than random-view selections.}

\noindent\textbf{Results on Open-vocabulary Queries}:
\textcolor{black}{To evaluate open-vocabulary querying performance, we design 6 to 8 natural language tasks per scene to locate specific functional elements on articulated objects. 
As presented in Tab. II, ArtiSG outperforms Fun3DU due to two fundamental methodological advantages.  
Firstly, our pipeline pre-constructs a comprehensive set of functional element nodes before performing the query. 
This ensures a highly complete candidate pool, naturally yielding a much higher recall (e.g., 77.4\% vs. 48.2\% in real scenes). 
Secondly, Fun3DU relies on prompting a VLM with templates like ``Point to all the [elements] in order to [task]'' to generate 2D points, which are subsequently fed into SAM for masking. 
Due to the inherent hallucination issues of VLMs in complex physical scenes, this paradigm frequently leads to imprecise localization or entirely incorrect region selections, severely degrading precision. 
In contrast, our node-level open-vocabulary features ensure robust and accurate semantic matching. 
These advantages are further demonstrated by the qualitative results in \Cref{fig:vis_artisg}.
}

\begin{table}[t]
\centering
\color{black}
\caption{Open-Vocabulary Query Evaluation}
\begin{tabular}{clccc}
\toprule
\textbf{Scene} & \textbf{Method} & $\mathbf{R}_{>0}$ & $\mathbf{P}_{>0}$ & $\mathbf{F1}_{>0}$ \\

\midrule
\multirow{2}{*}{Real-world} 
& Fun3DU\cite{corsetti2025fun3du} & 48.2 & 48.8 & 48.2 \\
& ArtiSG (ours)                & \bf{77.4} & \bf{59.5} & \bf{66.5} \\

\midrule
\multirow{2}{*}{Simulation} 
& Fun3DU\cite{corsetti2025fun3du} & 43.4 & 47.6 & 45.2 \\
& ArtiSG (w.o human)                    & \bf{68.9} & \bf{62.2} & \bf{64.9} \\

\bottomrule
\end{tabular}
\label{table:query-fun3du}
\end{table}

\subsection{Viewpoint-robust Articulation Tracking Evaluation}
\label{sec: eval_tracking}

\noindent\textbf{Setup}:
We evaluate our articulation tracking using an iPhone 12 Pro for camera pose estimation and a UMI gripper equipped with an ArUco sphere for end-effector tracking. 
We utilize an OptiTrack motion capture system to acquire ground-truth trajectories. 
The evaluation is conducted under two settings: a \textit{static view}, where the operator remains stationary to minimize ego-motion, and a \textit{dynamic view}, where the operator moves naturally during manipulation to introduce realistic camera jitter and viewpoint changes.

\begin{figure}[t]
    \centering
    \includegraphics[width=1\linewidth]{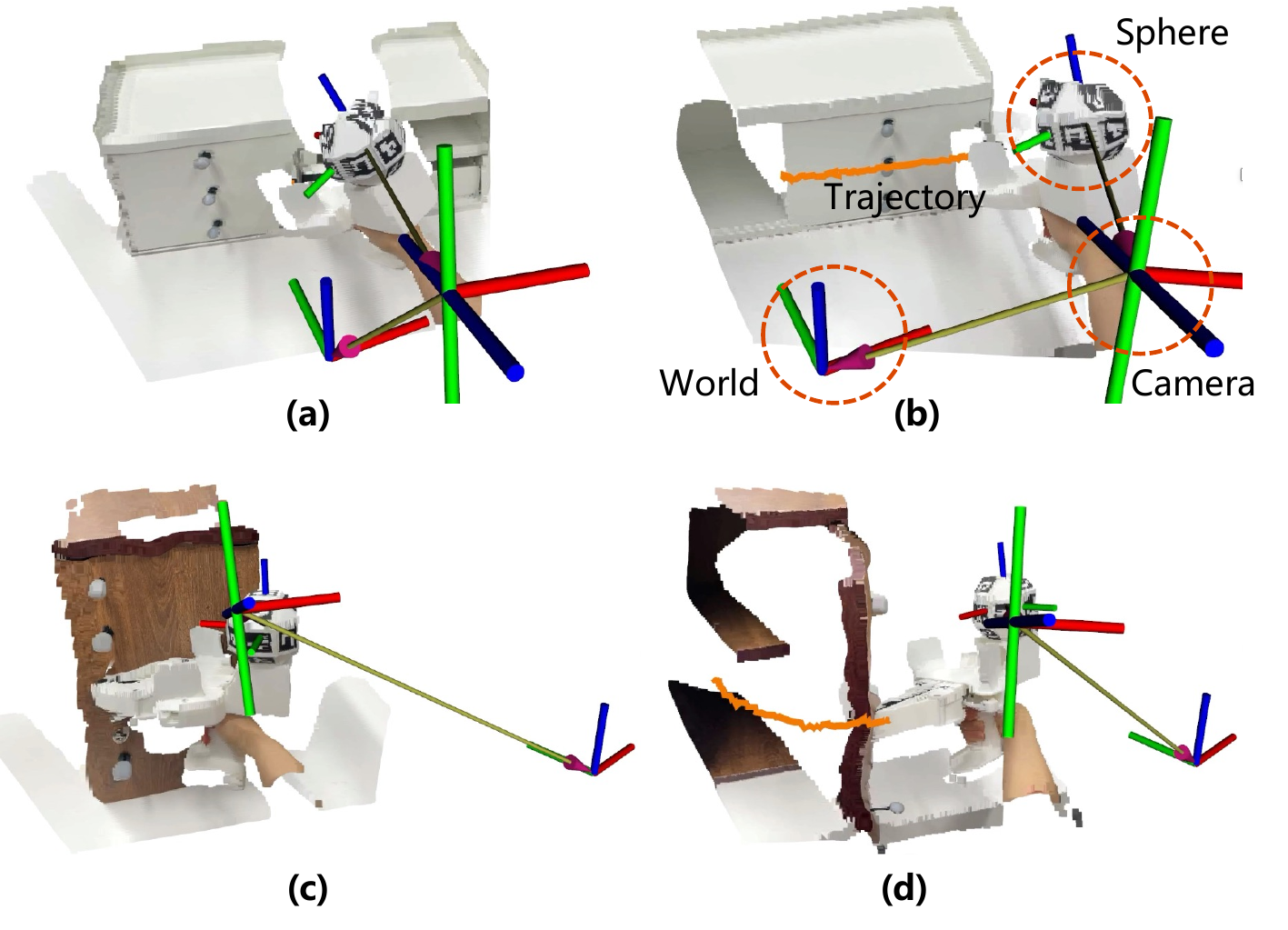}
    \setlength{\abovecaptionskip}{-0.4cm}
    \setlength{\belowcaptionskip}{-0.4cm}
    \caption{\textbf{Visualization of the viewpoint-robust articulation tracking process.} Subfigures (a) and (b) depict the start and end phases of manipulating a prismatic joint, while (c) and (d) show the manipulation of a revolute joint. The distinct coordinate frames for the World, Camera, and Sphere are highlighted to illustrate our decoupled tracking setup. The recovered gripper trajectory is visualized as an orange curve, demonstrating smooth and precise tracking performance.}
    \label{fig:tracking}
\end{figure}

\noindent\textbf{Baselines}: 
\textcolor{black}{We compare against four representative baselines: 
GFlow \cite{yuan2025general} for static image inference, 
CoTracker \cite{karaev24cotracker3} and ArtiPoint \cite{arti25werby} for tracking object surface points,
and Mediapipe \cite{lugaresi2019mediapipe} for hand tracking. 
Crucially, because our UMI gripper inherently occludes both the operator's hand and part of the object's surface, we evaluate baselines on separate bare-hand manipulation sequences to ensure a fair comparison. 
Furthermore, as GFlow, CoTracker, and Mediapipe output 3D motion trajectories, we directly substitute their extracted trajectories into our axis estimation module to compute the final articulation parameters.}

\noindent\textbf{Metrics}:
We calculate the trajectory RMSE $T_\mathrm{err}$ by measuring the Euclidean distance error between the estimated and ground truth trajectories.
The accuracy of the articulation axis estimation is also assessed by the axis angular error $\theta_\mathrm{err}$, which measures the deviation in the direction of the estimated axis, and the axis position error $d_\mathrm{err}$, which quantifies the distance deviation of the axis origin for revolute joints.

\setlength{\tabcolsep}{2pt}
\begin{table}[t!]
\centering
\caption{Quantitative comparison of articulation estimation performance.}
\label{tab:articulation_tracking}
\begin{tabular}{llccccc}
\toprule
\multirow{3}{*}{\textbf{Setting}} & \multirow{3}{*}{\textbf{Method}} & \multicolumn{2}{c}{\textbf{Prismatic joints}} & \multicolumn{3}{c}{\textbf{Revolute joints}} \\
\cmidrule(lr){3-4} \cmidrule(lr){5-7}
 & & \makecell{$T_\mathrm{err}$ \\ (cm) $\downarrow$} & \makecell{$\theta _\mathrm{err}$\\ (deg) $\downarrow$} &  \makecell{$T_\mathrm{err}$\\ (cm) $\downarrow$} & \makecell{$\theta _\mathrm{err}$\\ (deg) $\downarrow$} & \makecell{$d_\mathrm{err}$\\ (cm) $\downarrow$} \\
\midrule
\multirow{5}{*}{\makecell[c]{Static}}
& GFlow \cite{yuan2025general} & - & 16.610   &  - & 38.084  & 13.122  \\
\addlinespace[3pt]
& \textcolor{black}{ArtiPoint\cite{arti25werby}} & \textcolor{black}{-} & \textcolor{black}{16.334} & \textcolor{black}{-} & \textcolor{black}{11.731} & \textcolor{black}{22.700} \\
\addlinespace[3pt]
& CoTracker \cite{karaev24cotracker3} & 14.342 & 4.145 & 7.310 & 4.976 & 1.883\\
\addlinespace[3pt]
& Mediapipe \cite{lugaresi2019mediapipe} & 1.788 & 3.703   & 3.826 & 4.066 & 2.219 \\
\addlinespace[3pt]
& ArtiSG(ours) & \textbf{0.976} & \textbf{1.026}   & \textbf{1.092} & \textbf{1.627} & \textbf{0.811} \\
\midrule
\multirow{4}{*}{\makecell[c]{Dynamic}}
& \textcolor{black}{ArtiPoint\cite{arti25werby}} & \textcolor{black}{-} & \textcolor{black}{17.361} & \textcolor{black}{-} & \textcolor{black}{39.326} & \textcolor{black}{19.281} \\
\addlinespace[3pt]
& CoTracker \cite{karaev24cotracker3}  & 7.967 & 1.541 & 11.039 & 5.016  & 3.619  \\
\addlinespace[3pt]
& Mediapipe \cite{lugaresi2019mediapipe} & 1.083 & 2.291   & 2.953 & 6.644 & 3.757 \\
\addlinespace[3pt]
& ArtiSG(ours) & \textbf{0.820} & \textbf{1.314} & \textbf{0.899} & \textbf{2.322} & \textbf{1.225} \\
\bottomrule
\end{tabular}
\end{table}

\noindent\textbf{Results}:
To visually demonstrate the effectiveness of our pipeline, \Cref{fig:tracking} illustrates the coordinate frame transformations and the recovered trajectories for both prismatic and revolute joints during manipulation.
Quantitative comparisons are presented in \Cref{tab:articulation_tracking}. 
Overall, ArtiSG demonstrates superior accuracy and robustness compared to the above baselines.
\textbf{1) Comparison with Static Inference:} 
As shown in the static setting, GFlow struggles to accurately estimate articulation parameters, yielding high angular errors. This highlights the inherent ambiguity of inferring kinematics from static visual observations alone, validating our choice of using interaction trajectories.
\textcolor{black}{\textbf{2) Tracking Accuracy \& Hardware Trade-off:}}
\textcolor{black}{Compared to hardware-free baselines, our method significantly reduces tracking and estimation errors. 
For instance, while the concurrent work ArtiPoint successfully relaxes hardware constraints, it exhibits high angular and position errors. 
Similarly, compared to CoTracker and Mediapipe, we reduce the static revolute trajectory RMSE from 7.31 cm and 3.83 cm to just 1.09 cm.}
This improvement stems from our rigid-body tracking approach, which bypasses the jitter and surface contact variations common in hand tracking as well as the performance degradation due to textureless surfaces and occlusion in point tracking, enabling reliable kinematic inference.
\textbf{3) Robustness to Dynamics:}
Our pipeline maintains high accuracy even in a dynamic setting.
This robustness is attributed to our decoupled setup: the head-mounted camera tracks the markers on the UMI gripper, which ensures high-quality pose estimation regardless of the operator's body movement.
\textbf{4) Performance Across Joint Types:}
While all methods perform reasonably well on simpler prismatic joints, the advantage of ArtiSG is most pronounced on geometrically complex revolute joints. 
Accurately estimating the rotation axis requires precise arc fitting, which is sensitive to noise. 
Our method effectively handles this complexity, reducing the trajectory RMSE by approximately 70\% compared to the best-performing baseline Mediapipe in dynamic scenarios.

\begin{figure}[t]
    \centering
    \includegraphics[width=1\linewidth]{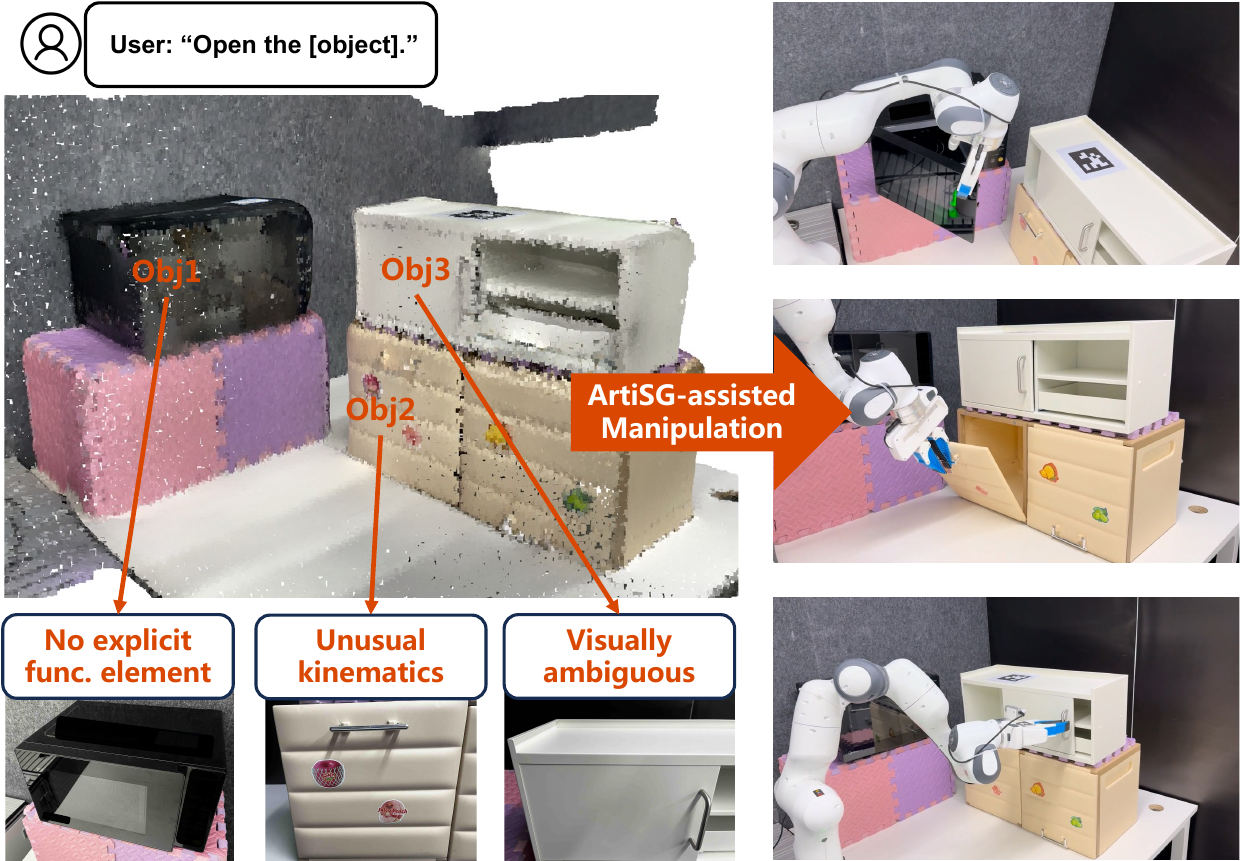}
    \setlength{\abovecaptionskip}{-0.4cm}
    \setlength{\belowcaptionskip}{-0.4cm}
    \caption{\textbf{Demonstration of ArtiSG-assisted robot manipulation.} We evaluate the system on objects with inconspicuous elements (Obj1), unusual kinematics (Obj2), and visual ambiguity (Obj3). While
    \textcolor{black}{inference from static observations} struggles to correctly infer the articulation mechanisms from appearance alone (e.g., mistaking the flip-down door of Obj2 for a drawer), ArtiSG-assisted method leverages the stored ``memory" of human demonstrations to retrieve precise 6-DoF end-effector trajectories, successfully executing the opening tasks (Right).}
    \label{fig:application}
\end{figure}

\subsection{ArtiSG-assisted Robot Manipulation \textcolor{black}{Demonstration}}
\label{sec:application}

To demonstrate the practical downstream utility of ArtiSG, we conduct real-world experiments using a Franka Research 3 robot arm. 
The robot is tasked with executing natural language commands ``Open the [object]'' on a set of challenging articulated objects shown in \Cref{fig:application}. 
These objects are specifically selected to highlight common perception difficulties, including \textbf{inconspicuous functional elements} (Obj1, a microwave without explicit functional elements), \textbf{unusual kinematics} (Obj2, a flip-down box resembling a drawer), and \textbf{visual ambiguity} (Obj3, a cabinet with unclear opening mechanisms).

\textcolor{black}{Relying solely on static observations to infer interactions for such deceptive objects is highly prone to failure, such as mistaking the flip-down door for a pull-out drawer.
In contrast, ArtiSG bypasses these perceptual ambiguities entirely by leveraging its stored kinematic memory of human demonstrations.} 
Upon receiving a command, the system queries the scene graph to retrieve the target functional node and its demonstrated 6-DoF trajectory. 
As visualized in \Cref{fig:application}, explicitly guiding the robot using these actionable kinematic priors ensures successful task execution. 
\textcolor{black}{This confirms that ArtiSG serves as a reliable robotic memory, effectively bridging the gap between semantic understanding and physically grounded manipulation.}

\section{Conclusion and Future Work}
\label{sec:con}
In this work, we introduced ArtiSG, \textcolor{black}{a systematic framework} that bridges the gap between semantic scene understanding and physical interaction by encoding human demonstrations into functional 3D scene graphs. 
By leveraging a viewpoint-robust tracking pipeline and an interaction-augmented refinement method, our system effectively resolves the visual ambiguities inherent in static perception and captures inconspicuous functional elements often missed by general detectors. 
\textcolor{black}{Nevertheless, our current data collection pipeline requires a gripper-based interface for motion tracking, making demonstration acquisition less natural than direct hand interaction. 
In future work, we plan to investigate markerless tracking and develop a more ergonomic interface that better conforms to natural human hand movements.}
Furthermore, we aim to integrate ArtiSG with general robot manipulation policies, utilizing the structured kinematic priors stored in our graph as explicit guidance to facilitate robust and efficient task execution.

\bibliographystyle{IEEEtran}
\bibliography{7_reference}






\end{document}